\documentclass[letterpaper, 10 pt, conference]{ieeeconf}  
\IEEEoverridecommandlockouts                     
\usepackage{comment}
\usepackage{cite}
\usepackage{amsmath,amssymb,amsfonts}
\usepackage{graphicx}
\usepackage{textcomp}
\usepackage{xcolor}
\usepackage{mathrsfs} 
\usepackage{comment}
\usepackage{subcaption}
\usepackage[linesnumbered,ruled,vlined]{algorithm2e}
\usepackage[noend]{algpseudocode}
\usepackage{multirow}
\usepackage{nicematrix}
\usepackage{hhline}
\usepackage{hologo}
\usepackage{listings}
\usepackage{scalerel}
\usepackage{tikz}
\usetikzlibrary{svg.path}

\definecolor{orcidlogocol}{HTML}{A6CE39}
\tikzset{
	orcidlogo/.pic={
		\fill[orcidlogocol] svg{M256,128c0,70.7-57.3,128-128,128C57.3,256,0,198.7,0,128C0,57.3,57.3,0,128,0C198.7,0,256,57.3,256,128z};
		\fill[white] svg{M86.3,186.2H70.9V79.1h15.4v48.4V186.2z}
		svg{M108.9,79.1h41.6c39.6,0,57,28.3,57,53.6c0,27.5-21.5,53.6-56.8,53.6h-41.8V79.1z M124.3,172.4h24.5c34.9,0,42.9-26.5,42.9-39.7c0-21.5-13.7-39.7-43.7-39.7h-23.7V172.4z}
		svg{M88.7,56.8c0,5.5-4.5,10.1-10.1,10.1c-5.6,0-10.1-4.6-10.1-10.1c0-5.6,4.5-10.1,10.1-10.1C84.2,46.7,88.7,51.3,88.7,56.8z};
	}
}

\newcommand\orcidicon[1]{\href{https://orcid.org/#1}{\mbox{\scalerel*{
				\begin{tikzpicture}[yscale=-1,transform shape]
					\pic{orcidlogo};
				\end{tikzpicture}
			}{|}}}}

\usepackage{hyperref} 
\hypersetup{
	colorlinks=true,
	breaklinks=true,
	linkcolor=blue,
	filecolor=magenta,      
	hypertexnames=true,
	urlcolor=cyan,
	pdfpagemode=FullScreen,
	citecolor=green
}
\usepackage{cleveref}

\usepackage{afterpage}
\usepackage{array}
\newcolumntype{L}[1]{>{\raggedright\let\newline\\\arraybackslash\hspace{0pt}}m{#1}}
\newcolumntype{C}[1]{>{\centering\let\newline\\\arraybackslash\hspace{0pt}}m{#1}}
\newcolumntype{R}[1]{>{\raggedleft\let\newline\\\arraybackslash\hspace{0pt}}m{#1}}

\title{\LARGE \textbf{Sampling for Model Predictive Trajectory Planning in\\Autonomous Driving using Normalizing Flows}$^{*}$
	
\author{Georg Rabenstein\textsuperscript{\orcidicon{0009-0008-4213-0619}}, Lars Ullrich\textsuperscript{\orcidicon{0009-0001-8166-3118}},~\IEEEmembership{Graduate Student Member,~IEEE}, \\and Knut Graichen\textsuperscript{\orcidicon{0000-0003-2865-8093}},~\IEEEmembership{Senior Member,~IEEE}
\thanks{*This research is accomplished within the project ”AUTOtechagil” (FKZ 01IS22088Y). We acknowledge the financial support for the project by the Federal Ministry of Education and Research of Germany (BMBF).}
\thanks{The authors are with the Chair of Automatic Control, Friedrich-Alexander-Universität Erlangen-Nürnberg (FAU), Germany {\tt\footnotesize \{georg.rabenstein, lars.ullrich, knut.graichen\}@fau.de}}%
}}

\usepackage{setspace}
\begin{document}
\twocolumn[
\begin{@twocolumnfalse}
	\Huge {IEEE copyright notice} \\ \\
	\large {\copyright\ 2024 IEEE. Personal use of this material is permitted. Permission from IEEE must be obtained for all other uses, in any current or future media, including reprinting/republishing this material for advertising or promotional purposes, creating new collective works, for resale or redistribution to servers or lists, or reuse of any copyrighted component of this work in other works.} \\ \\
	
	{\Large Published in \emph{2024 35th IEEE Intelligent Vehicles Symposium (IV)}, Jeju Island, Korea, June 2 - 5, 2024.} \\ \\
	
	Cite as:
	
	\vspace{0.1cm}
	\noindent\fbox{%
		\parbox{\textwidth}{%
			G.~Rabenstein, L.~Ullrich, and K.~Graichen, ``Sampling for Model Predictive Trajectory Planning in\\Autonomous Driving using Normalizing Flows,''
			in \emph{2024 35th IEEE Intelligent Vehicles Symposium (IV)}, Jeju Island, Korea, 2024, pp. 2091-2096, doi: 10.1109/IV55156.2024.10588765.
		}%
	}
	\vspace{2cm}
	
\end{@twocolumnfalse}
]

\noindent\begin{minipage}{\textwidth}
	
\hologo{BibTeX}:
\footnotesize
\begin{lstlisting}[frame=single]
@inproceedings{rabenstein2024sampling,
	author={Rabenstein, Georg and Ullrich, Lars and Graichen, Knut},
	booktitle={2024 35th IEEE Intelligent Vehicles Symposium (IV)},
	title={Sampling for Model Predictive Trajectory Planning in Autonomous Driving 
		using Normalizing Flows},
	address={Jeju Island, Korea},
	year={2024},
	pages={2091-2096},
	doi={10.1109/IV55156.2024.10588765},
	publisher={IEEE}
}
\end{lstlisting}
\end{minipage}
\setcounter{page}{0}

\maketitle

\thispagestyle{plain}
\pagestyle{plain}

\begin{abstract}

Alongside optimization-based planners, sampling-based approaches are often used in trajectory planning for autonomous driving due to their simplicity. Model predictive path integral control is a framework that builds upon optimization principles while incorporating stochastic sampling of input trajectories. This paper investigates several sampling approaches for trajectory generation. In this context, normalizing flows originating from the field of variational inference are considered for the generation of sampling distributions, as they model transformations of simple to more complex distributions. Accordingly, learning-based normalizing flow models are trained for a more efficient exploration of the input domain for the task at hand. The developed algorithm and the proposed sampling distributions are evaluated in two simulation scenarios.
\end{abstract}

\section{Introduction}

Trajectory planning is a relevant task in the field of autonomous driving and deals with the execution of local driving maneuvers while considering dynamic changes in the environment. Thereby, multiple objectives such as efficiency, safety, driver comfort and adhering to the road limits have to be considered \cite{katrakazas15}. For this purpose, optimization based model predictive control (MPC) approaches which consider the future state of the environment can be utilized \cite{betz22}. Thereby, it is possible to directly consider constraints such as the vehicle dynamics in the planning process.

Besides local gradient methods \cite{claussmann20}, sampling-based approaches \cite{gonzalez16, paden16} can be used to solve the underlying optimization problem. In particular, the gradient-free nature of those approaches provides high flexibility in terms of the design of cost and constraint functions \cite{sacks23}. For instance, sampling-based model predictive path integral control (MPPI) was deployed in the autonomous driving context \cite{williams16, williams18}. This algorithm is theoretically based on importance sampling \cite{botev13} and theoretic quantities like the free energy and the Kullback-Leibler (KL) divergence \cite{theodorou12}. The underlying optimization problem is hereby solved by stochastic sampling of input trajectories that are forward integrated via a vehicle dynamics model.

The choice of the sampling distribution can have a large impact on the performance of the algorithm, as it directly affects the resulting shape of the generated trajectories. Due to their simplicity, basic Gaussians are often used for sampling in the context of trajectory optimization \cite{williams16, williams18}. \\ Such an approach may however lead to a suboptimal performance, as a large percentage of high-cost trajectories may be generated \cite{sacks23}. Consequently, the algorithm is more likely to get stuck in local optima which increases the computational load. 

In contrast, normalizing flows \cite{rezende15} are a frequently used tool to improve the sampling distribution due to their ability to learn complex target distributions. For instance, \cite{sacks23}, \cite{agarwal20}, \cite{power22}, \cite{power24} and \cite{ma20} deploy normalizing flows in planning tasks. Those approaches motivate us to analyze normalizing flows in the context of trajectory sampling in model predictive trajectory planning.

In \cite{sacks23}, the authors deploy normalizing flows in a MPC setup and learn a sampling distribution for a low-sample regime in a bilevel optimization. While high performance in the learned regime is demonstrated, the distribution is not trained for multiple environments. In contrast, this work aims to build normalizing flow based sampling distributions that can be deployed in different environments. 

In the approach in \cite{agarwal20}, expert trajectories are used to train a conditional normalizing flow for imitative planning for a trajectory planning task in autonomous driving. Hereby, a variational autoencoder is first trained to extract features as a lower-rank representation of expert trajectories. A normalizing flow is then trained on these features.

The authors of \cite{power22} and \cite{power24} further demonstrate the applicability of normalizing flows in the context of sampling-based MPC. In \cite{power22}, the authors propose FlowMPPI for the collision-free navigation of robots and extend their methodology to iCEM in \cite{power24}. In both works, normalizing flows are conditioned on control inputs that include information on start, goal and environment. By using a projection method, the authors demonstrate the applicability of their methodology to out-of-distribution environments. In contrast to \cite{agarwal20}, \cite{power22} and \cite{power24}, training trajectories in our work do not include explicit information of the environment.
Our approach is similar to \cite{ma20} where normalizing flows are applied in vehicle trajectory forecasting to learn a more diverse sampling distribution. Hereby, an additional objective is introduced in the training process that encourages trajectories with high spatial separation.

The contribution of this paper is to use normalizing flows to learn sampling distributions that can explore the input regime more efficiently than existing basic approaches and that can be deployed in different driving scenarios. In contrast to introducing an additional training objective in \cite{ma20}, simple heuristic rules are introduced for the generation of training samples in this work. Existing sampling approaches are used to generate trajectories that are then altered by these rules. The altered training trajectories then form a distribution which is learned by a normalizing flow. After the training, the learned distributions are deployed for stochastic sampling in a MPPI based trajectory planning algorithm. The performance of normalizing flow models is compared to simple sampling distributions. For the evaluation, two scenarios with urban-like driving environments are created in Python. 

The paper is structured as follows: Section \ref{sec:mppi} describes the fundamentals of MPPI based trajectory planning. The normalizing flow based sampling approaches are presented in Section \ref{sec:nfs}. Section \ref{sec:results} presents simulation results to evaluate the performance of the algorithm. This paper ends with a conclusion in Section \ref{sec:conclusion}.
\section{State-of-the-art Model predictive path integral control}
\label{sec:mppi}

In the following, MPPI is described as a sampling-based trajectory planning method. The first subsection presents the fundamentals while the successive subsections deal with sampling approaches for trajectory generation.

\subsection{Fundamentals}

MPPI is a stochastic optimal control method which was introduced in \cite{williams16, williams18}. In this method, an optimization problem is solved over a finite time horizon by the iterative sampling of input trajectories. Consider a discrete system
\begin{align}
    \boldsymbol{x}_{i+1} = \boldsymbol{F}(\boldsymbol{x}_i,\boldsymbol{u}_i + \boldsymbol{v}_i)
    \label{eq:dynamics}
\end{align}
with $\boldsymbol{x}_i \in \mathbb{R}^{m_x}$ and $\boldsymbol{u}_i \in \mathbb{R}^{m_u}$ as system state and input. A random variable $\boldsymbol{v}_i  \sim \mathcal{N}(\boldsymbol{0}, \boldsymbol{\Sigma})$ is applied to the input which is drawn from a Gaussian with zero mean covariance $\boldsymbol{\Sigma}$. Hereby, $X\!=\![\boldsymbol{x}_1, \boldsymbol{x}_2, ...,\boldsymbol{x}_{N}] \! \in \! \mathbb{R}^{m_x \times N}$ and $U=[\boldsymbol{u}_0, \boldsymbol{u}_1, ..., \boldsymbol{u}_{N-1}] \in \mathbb{R}^{m_u \times N}$ are defined as state and input trajectories. Further, $V=[\boldsymbol{v}_0, \boldsymbol{v}_1,..., \boldsymbol{v}_{N-1}] \in \mathbb{R}^{m_u \times N}$ is introduced as a random trajectory. For trajectory planning, an optimization problem
\begin{align}
    \underset{X,U}{\text{argmin}} \,\, \mathbb{E}_{\mathbb{Q}}\bigl[ \underbrace{\phi(\boldsymbol{x}_N) + \sum_{i=0}^{N-1} \bigl(c(\boldsymbol{x}_i) + \frac{1}{2} \boldsymbol{u}_i^\top \boldsymbol{R} \boldsymbol{u}_i \bigr)}_{S(X,U)}\bigr]
    \label{eq:MPC_problem}
\end{align}
is defined where $\phi(\boldsymbol{x}_N)$, $c(\boldsymbol{x}_i)$ and $\boldsymbol{R} \in \mathbb{R}^{m_u \times m_u}$ are the terminal cost, the state dependent running cost and the positive definite control weight matrix. The total trajectory costs are denoted as $S(X,U)$. 
In order to compute the optimal trajectories $X^*, U^*$, $K$ random trajectories are drawn from a sampling distribution $\mathbb{Q}$. A sampled trajectory is hereby denoted as $V^{(k)}$. By combining $V^{(k)}$ with the mean input trajectory $\bar{U}$, the corresponding input trajectory $U^{(k)}$ with 
\begin{align}
    \boldsymbol{u}_i^{(k)} \leftarrow \bar{\boldsymbol{u}} + \boldsymbol{v}_i^{(k)} \,\, \forall i \in \{0, N-1\}
\end{align}
can be acquired. In our case, the solution from the previous algorithm step is used for $\bar{U}$. Initially, $\bar{U}\leftarrow \boldsymbol{0}$ is set.
The forward integration of $U^{(k)}$ via \eqref{eq:dynamics} then produces the state trajectory $X^{(k)}$. The optimal input trajectories are determined by a control law which is based on minimizing the KL divergence between $\mathbb{Q}$ and the latent optimal distribution $\mathbb{Q}^*$ \cite{williams18}. A weight factor 
\begin{align}
    w^{(k)} = \text{exp}(-\frac{1}{\lambda}S^{(k)})) 
    \label{eq:weight_determination}
\end{align}
is assigned to every trajectory where $\lambda \in \mathbb{R}^+$ is the inverse temperature. An iterative averaging scheme then updates
\begin{align}
    \boldsymbol{u}_i^* \gets \boldsymbol{u}_i^* + \frac{\sum_{k=1}^K w^{(k)}{\boldsymbol{u}_i^{(k)}}}{\sum_{k=1}^K w^{(k)}}
\end{align}
towards the optimal input trajectory. Besides the design of suitable cost functions, the choice of a sampling distribution $\mathbb{Q}$ can have a considerable impact on the performance of the algorithm. The sampling process deals with the generation of random trajectories $V^{(k)}$ which are then applied to the input. Due to their simplicity, basic Gaussians (BG) are often used for sampling \cite{williams16, williams18}. Hereby, every random vector
\begin{align}
    \boldsymbol{v}_i \sim \mathcal{N}(\boldsymbol{0}, \boldsymbol{\Sigma}) \,\, \forall i \in \{0, N-1\}
    \label{eq:basic_gaussian}
\end{align}
is directly drawn from a zero-mean Gaussian. Input trajectories can however be restricted in their shape. This may result in an increased number of high-cost samples and can limit the performance of the algorithm \cite{sacks23}. Sampling from a BG distribution can also produce significant chattering of the input and state trajectories \cite{williams18, kim22}. Two alternative sampling approaches are described in the two following subsections.

\subsection{Input lifting}
\label{sec:il}

Input lifting (IL) sampling is inspired by \cite{kim22}. In this regard, a trajectory $\dot{V} = [\dot{\boldsymbol{v}}_0, \dot{\boldsymbol{v}}_1, ..., \dot{\boldsymbol{v}}_{N-1}] \in \mathbb{R}^{m_u \times N}$ is drawn on a derivative level from a zero-mean Gaussian \eqref{eq:basic_gaussian}. Compared to the BG approach, a discrete integration step is added in the sampling process. The trajectory $V$ is then computed via
\begin{align}
    \boldsymbol{v}_i = \boldsymbol{v}_{i-1} + \dot{\boldsymbol{v}}_{i-1} \Delta t \,\, \forall i \in \{1, N-1\}
\end{align}
with $\boldsymbol{v}_0=\boldsymbol{0}$ and the discretization time $\Delta t$. The discrete integration step produces input trajectories with higher smoothness.

\subsection{Two-degree-of-freedom sampling}
\label{sec:2df}

This approach combines BG and IL sampling and is therefore referred to as two-degree-of-freedom (2DF) sampling. Two sequences of random vectors $\Tilde{V}^{(1)}, \Tilde{V}^{(2)} \in \mathbb{R}^{m_u \times N}$ with
\begin{align}
    \Tilde{\boldsymbol{v}}_i^{(j)} \sim \mathcal{N}(\boldsymbol{0},\boldsymbol{\Sigma}^{(j)}) \quad \forall i \in \{0, N-1\}, \, j \in \{1,2\}
\end{align}
are drawn from zero-mean Gaussians with covariances $\boldsymbol{\Sigma}^{(1)}$ and $\boldsymbol{\Sigma}^{(2)}$. Note that $\Tilde{V}^{(1)}$ are drawn on a derivative level. The trajectory $V$ is then computed by the discrete integration of $\Tilde{V}^{(1)}$ and the addition of $\Tilde{V}^{(2)}$, i.e. 
\begin{subequations}
\begin{align}
    \boldsymbol{v}_0 &= \boldsymbol{0} \\
    \boldsymbol{v}_i &= \boldsymbol{v}_{i-1} + \Tilde{\boldsymbol{v}}_{i-1}^{(1)} \Delta t \,\,\,\, \forall i \in \{1, N-1\} \\
    \boldsymbol{v}_i &\leftarrow \boldsymbol{v}_i + \Tilde{\boldsymbol{v}}_i^{(2)} \quad \quad \,\,\,\,\,\, \forall i \in \{0, N-1\}. 
\end{align}
\label{eq:2df_computation}
\end{subequations}
The use of two separate Gaussians with $\boldsymbol{\Sigma}^{(1)}$ and $\boldsymbol{\Sigma}^{(2)}$ can potentially lead to a better exploration of the input domain.

\section{Normalizing flow based sampling approaches}
\label{sec:nfs}

Due to their ability to model diverse probability distributions, normalizing flows can be advantageous for stochastic trajectory sampling. Two normalizing flow based sampling concepts are introduced in this section. Trajectories that are generated by the previously described sampling approaches are hereby altered via probabilistic rules. The resulting trajectories then form a distribution that is learned by a normalizing flow. The fundamentals of normalizing flows are presented in the following. Afterwards, the sample generation is described and details on the implementation are given.

\subsection{Fundamentals}

Due to their ability to estimate and build arbitrarily complex probability distributions, normalizing flows are a suitable tool for variational inference \cite{rezende15, tabak13}. Such probabilistic models are theoretically based on the change of variable formula. Hereby, consider a random variable $\boldsymbol{z} \in \mathbb{R}^D$ with the known probability density $q(\boldsymbol{z})$. The dimensionality-preserving mapping $\boldsymbol{g}:\mathbb{R}^D \rightarrow \mathbb{R}^D$ can then be used to transform $\boldsymbol{z}$ to $\boldsymbol{z}'=\boldsymbol{g}(\boldsymbol{z})$. The corresponding density function results as
\begin{align}
    q(\boldsymbol{z}') = q(\boldsymbol{z}) \Bigl| \text{det} \frac{\partial \boldsymbol{g}^{-1}}{\partial \boldsymbol{z}'}\Bigr|=q(\boldsymbol{z})\Bigl| \text{det} \frac{\partial \boldsymbol{g}}{\partial \boldsymbol{z}}\Bigr|^{-1}.
\end{align}
A normalizing flow, which is parameterized by $\Theta$, applies a sequence of such transformations to the known base distribution $p(\boldsymbol{z})$ and has the resulting posterior $q_\Theta (\boldsymbol{z}')$. The goal hereby is to find a good approximation of the latent posterior $p(\boldsymbol{z}')$. This is achieved by minimizing the KL divergence between $q_\Theta (\boldsymbol{z}')$ and $p(\boldsymbol{z}')$
\begin{align}
    &\mathbb{D}_{KL}(q_\Theta(\boldsymbol{z}')||p(\boldsymbol{z}')) = \int q_\Theta(\boldsymbol{z}') \text{log} \frac{q_\Theta(\boldsymbol{z}')}{p(\boldsymbol{z}')} d\boldsymbol{z} \nonumber \\
    &= \mathbb{E}_{q_\Theta(\boldsymbol{z}')}[\text{log} \, q_\Theta(\boldsymbol{z}') - \text{log} \, p(\boldsymbol{z}')] \\
    &= \mathbb{E}_{p(\boldsymbol{z})} [\text{log} \, p(\boldsymbol{z}) - \text{log} \Bigl| \text{det} \frac{\partial \Theta}{\partial \boldsymbol{z}}\Bigr| - \text{log} \, p(\boldsymbol{z}')] \nonumber
\end{align}
by adjusting the flow parameters $\Theta$ \cite{power22}. Formally, a normalizing flow consists of a series of $L$ transformations which are successively applied to an input variable $\boldsymbol{z}_0$ with the density $q_0(\boldsymbol{z}_0)$. This produces an output variable
\begin{align}
    \boldsymbol{z}_L=\boldsymbol{g}_L \circ ... \circ \boldsymbol{g}_2 \circ \boldsymbol{g}_1(\boldsymbol{z}_0)
\end{align}
which has the probability density
\begin{align}
    \text{log} \, q_L(\boldsymbol{z}_L) = \text{log} \, q_0(\boldsymbol{z}_0) - \sum_{l=1}^{L} \text{log} \, \text{det} \Bigl| \frac{\partial \boldsymbol{g}_l}{\partial \boldsymbol{z}_l}\Bigr|.
\end{align}

\subsection{Normalizing flow based two-degree-of-freedom sampling}
\label{sec:nf_a2DF}

\begin{algorithm}[b]
\DontPrintSemicolon

  \small
  $V_{b, i}^{(j)} \sim \mathcal{N}(0, \epsilon_{\mathrm{draw}}^{(j)}) \,\,\,\, \forall b \in \{1, B\}, \, i \in \{1, N\}, \, j \in \{1,2\}$ \\
  $\rho_b^{(j)} = \sum_{i=1}^N V_{b,i}^{(j)} \quad \,\,\,\forall b \in \{1, B\}, \, j \in \{1,2\}$ \\
  $\hat{\boldsymbol{V}}^{(1)} = \mathrm{ascentSort}(\mathrm{array}=\boldsymbol{V}^{(1)}, \mathrm{measure}=\boldsymbol{\rho}^{(1)})$ \\
    $\hat{\boldsymbol{V}}^{(2)} = \mathrm{descentSort}(\mathrm{array}=\boldsymbol{V}^{(2)}, \mathrm{measure}=\boldsymbol{\rho}^{(2)})$ \\ 

    \For{$b=1:B$}{
        $\hat{b}_1, \hat{b}_2 = \mathrm{drawViaHeuristic}()$ \tcp*{Eq. \eqref{eq:heuristic}}
        $\boldsymbol{V}_b^{*}=\mathrm{joinTrajectories}(\hat{\boldsymbol{V}}_{\hat{b}_1}^{(1)}, \hat{\boldsymbol{V}}_{\hat{b}_2}^{(2)})$ \tcp*{Eq. \eqref{eq:2df_computation}}
    }
\caption{Trajectory generation for normalizing flow based adaptive two-degree-of-freedom sampling}
\label{alg:samples_nf_adof}
\end{algorithm}

The normalizing flow based adaptive two-degree-of-freedom (NF-A2DF) sampling concept is described in this subsection. The approach from Sec. \ref{sec:2df} is adapted by a heuristic rule to improve the sampling efficiency. This rule is kept very general and increases the chance of directional changes within a trajectory. 
Algorithm \ref{alg:samples_nf_adof} describes the generation of training trajectories $\boldsymbol{V}^* \in \mathbb{R}^{B \times N}$ which are then used to train a normalizing flow. Again, $N$ marks the trajectory length while the total number of samples is denoted by $B$. In this work, the dynamics model features two control inputs with different magnitudes. For that reason, separated flow models are trained for every input which results in $m_u=1$. 
Two groups of samples $\boldsymbol{V}^{(1)}, \boldsymbol{V}^{(2)} \in \mathbb{R}^{B \times N}$ are drawn from zero-mean Gaussians with variance $\epsilon_{\mathrm{draw}}^{(1)}$ and $\epsilon_{\mathrm{draw}}^{(2)}$. The samples from both groups are assessed via the trajectory sums $\boldsymbol{\rho}^{(1)}, \boldsymbol{\rho}^{(2)} \in \mathbb{R}^B$. These are acquired by successively computing the sum of all entries over the full trajectory length $N$. The samples are then sorted via $\boldsymbol{\rho}^{(1)}, \boldsymbol{\rho}^{(2)}$ which results in $\hat{\boldsymbol{V}}^{(1)}, \hat{\boldsymbol{V}}^{(2)} \in \mathbb{R}^{B \times N}$. The trajectories from the first group are sorted in ascending and the trajectories from the second group in descending order. Trajectories from both groups are combined via a heuristic rule 
\begin{align}
    &\hat{b}_1 \sim \mathrm{drawUniformIndex}(\mathrm{min}=1, \mathrm{max}=B) \nonumber \\
    &\hat{b}_2 \sim \mathcal{N}(\Tilde{b}_1, \epsilon_{\mathrm{switch}}) \label{eq:heuristic} \\
    &\hat{b}_2 \leftarrow \mathrm{clip}( \lceil \hat{b}_2 \rceil, \mathrm{min}=1, \mathrm{max}=B). \nonumber
\end{align}
Hereby, a trajectory number $\hat{b}_1$ for the first group is drawn uniformly random. A Gaussian with mean $\hat{b}_1$ and variance $\epsilon_{\mathrm{switch}}$ is then used to generate the number $\hat{b}_2$ for the second group. Note that the last line in \eqref{eq:heuristic} ensures that $\hat{b}_2$ is an integer. Consider the case that $\hat{b}_1$ corresponds to a trajectory $\hat{\boldsymbol{V}}_{\hat{b}_1}^{(1)}$ that has a high trajectory sum. The complementary trajectory $\hat{\boldsymbol{V}}_{\hat{b}_1}^{(2)}$ from the second group would then have a low sum. This heuristic thus increases the chance to combine trajectories with different directions. Drawing with the variance $\epsilon_{\mathrm{switch}}$ introduces stochasticity in the trajectory generation. The combined trajectories $\boldsymbol{V}^{*}$ are computed via \eqref{eq:2df_computation} and define a sampling distribution which is learned by a normalizing flow. 

\subsection{Normalizing flow based adaptive input lifting}

\begin{algorithm}[tb]
\label{alg:samples_nf_ail}
 \caption{Trajectory generation for normalizing flow based adaptive input lifting}
 \small
$\dot{V}_{b,i}^{(j)} \sim \mathcal{N}(0, \epsilon_{\mathrm{draw}}) \,\, \forall b \in \{1, B\}, \, i \in \{1, \frac{N}{4}\}, \, j \in \{1, 4\}$ \\
 $\dot{\boldsymbol{V}}^* \leftarrow \mathrm{joinTrajectories}(\dot{\boldsymbol{V}}^{(1)}, \dot{\boldsymbol{V}}^{(2)})$ \\
 $\dot{\boldsymbol{V}}^* \leftarrow \mathrm{joinTrajectories}(\dot{\boldsymbol{V}}^*, \dot{\boldsymbol{V}}^{(3)})$ \\
 $\dot{\boldsymbol{V}}^* \leftarrow \mathrm{joinTrajectories}(\dot{\boldsymbol{V}}^*, \dot{\boldsymbol{V}}^{(4)})$ \\

\SetKwInput{KwInput}{Input}                
\SetKwInput{KwOutput}{Output}              
\DontPrintSemicolon
  
  \KwInput{\,\,\,\,$\tilde{\boldsymbol{V}}^{(1)} \in \mathbb{R}^{\Tilde{B} \times \Tilde{N}_1}$, $\tilde{\boldsymbol{V}}^{(2)} \in \mathbb{R}^{\Tilde{B} \times \Tilde{N}_2}$}
  \KwOutput{$\tilde{\boldsymbol{V}}^{*} \in \mathbb{R}^{\Tilde{B} \times (\Tilde{N}_1 + \Tilde{N}_2)}$}

  \SetKwFunction{FMain}{generateTrajectories}
  \SetKwFunction{FJoin}{joinTrajectories}
 
  \SetKwProg{Fn}{Function}{:}{}
  \Fn{\FJoin{$\tilde{\boldsymbol{V}}^{(1)}$, $\tilde{\boldsymbol{V}}^{(2)}$}}{
  $\rho_{\Tilde{b}}^{(j)} = \sum_{i=1}^{\Tilde{N}_j} \Tilde{V}_{\Tilde{b},i}^{(j)} \quad \forall \Tilde{b} \in \{1, \Tilde{B}\}, \, j \in \{1, 2\}$ \\
  $\hat{\boldsymbol{V}}^{(1)} = \mathrm{ascentSort}(\mathrm{array}=\Tilde{\boldsymbol{V}}^{(1)}, \mathrm{measure}=\boldsymbol{\rho}^{(1)})$ \\
    $\hat{\boldsymbol{V}}^{(2)} = \mathrm{descentSort}(\mathrm{array}=\Tilde{\boldsymbol{V}}^{(2)}, \mathrm{measure}=\boldsymbol{\rho}^{(2)})$ \\ 

  \For{$\Tilde{b}=1:\Tilde{B}$}{
        $\hat{b}_1, \hat{b}_2 = \mathrm{drawViaHeuristic}()$ \tcp*{Eq. \eqref{eq:heuristic}}
        $\Tilde{\boldsymbol{V}}_{\Tilde{b}}^* = [\hat{\boldsymbol{V}}_{\hat{b}_1}^{(1)}, \hat{\boldsymbol{V}}_{\hat{b}_2}^{(2)}] $
        }
        \KwRet $\Tilde{\boldsymbol{V}}_{\Tilde{b}}^{(12)}$ \\
  }
\label{alg:samples_nf_ail} 
\end{algorithm}

The sampling concept from Sec. \ref{sec:il} is also adapted via the heuristic rule from \eqref{eq:heuristic} to improve the sampling efficiency. We refer to this approach as normalizing flow based adaptive input lifting (NF-AIL). The generation of training trajectories is described in Alg. 2. The resulting training trajectories $\dot{\boldsymbol{V}}^* \in \mathbb{R}^{B \times N}$ are generated by combining trajectories
\begin{align}
    \dot{\boldsymbol{V}}^{(j)} \in \mathbb{R}^{B \times \frac{N}{4}} \quad \forall j \in \{1,4\}
\end{align}
from four segments where each trajectory has length $\frac{N}{4}$. Note that $N$ should be divisible by four. A zero-mean Gaussian with variance $\epsilon_{\mathrm{draw}}$ is used for sample generation. 
The trajectories $\dot{\boldsymbol{V}}^*$ are the result of three consecutive joining operations. Similar to the NF-A2DF concept, trajectories from two respective groups are sorted in ascending and descending order and the heuristic rule from \eqref{eq:heuristic} is employed to join two trajectories. In contrast to Sec. \ref{sec:nf_a2DF}, the trajectory length of $\dot{\boldsymbol{V}}^*$ increases with every joining operation. In this case, it is thus possible to combine trajectories with different length ($\Tilde{N}_1=\Tilde{N}_2$ is not required). 
 The trajectories $\dot{\boldsymbol{V}}^*$ form a distribution which is then learned by a normalizing flow. Sampling from the learned distribution then produces input trajectories with high smoothness and an increased chance of directional changes.

\subsection{Implementation}
\label{sec:nf_training}

The MPPI based trajectory planning algorithm is implemented for evaluation in Python. Hereby, the package \textit{Normflows} \cite{stimper23} is used for the practical implementation of normalizing flows. As a basic architecture, Residual Flow \cite{chen19} with $16$ residual flow layers is selected for every trained model. The trajectory length is selected as $N=80$ such that each flow layer also features a residual network $\boldsymbol{\Phi}$ with $N=80$ inputs. A network further has $128$ hidden units and $2$ hidden layers and a Lipschitz constant of $\mathrm{Lip}(\boldsymbol{\Phi})=0.9$. The network parameters are learned via a gradient descent algorithm which uses a marginal likelihood based loss function \cite{rezende15}. A zero-mean Gaussian with unit covariance is chosen as a prior distribution in every case. 
\section{Results}
\label{sec:results}

Two evaluation scenarios are created and the performance of the algorithm is observed for every sampling distribution. General implementation details are given in the first subsection. Afterwards, the training process for the normalizing flow based sampling concepts is described. The following subsections show the simulation results for both scenarios.

\subsection{Setup}
\label{sec:results_setup}

\begin{figure}[tb]
    \includegraphics[width=0.46\textwidth]{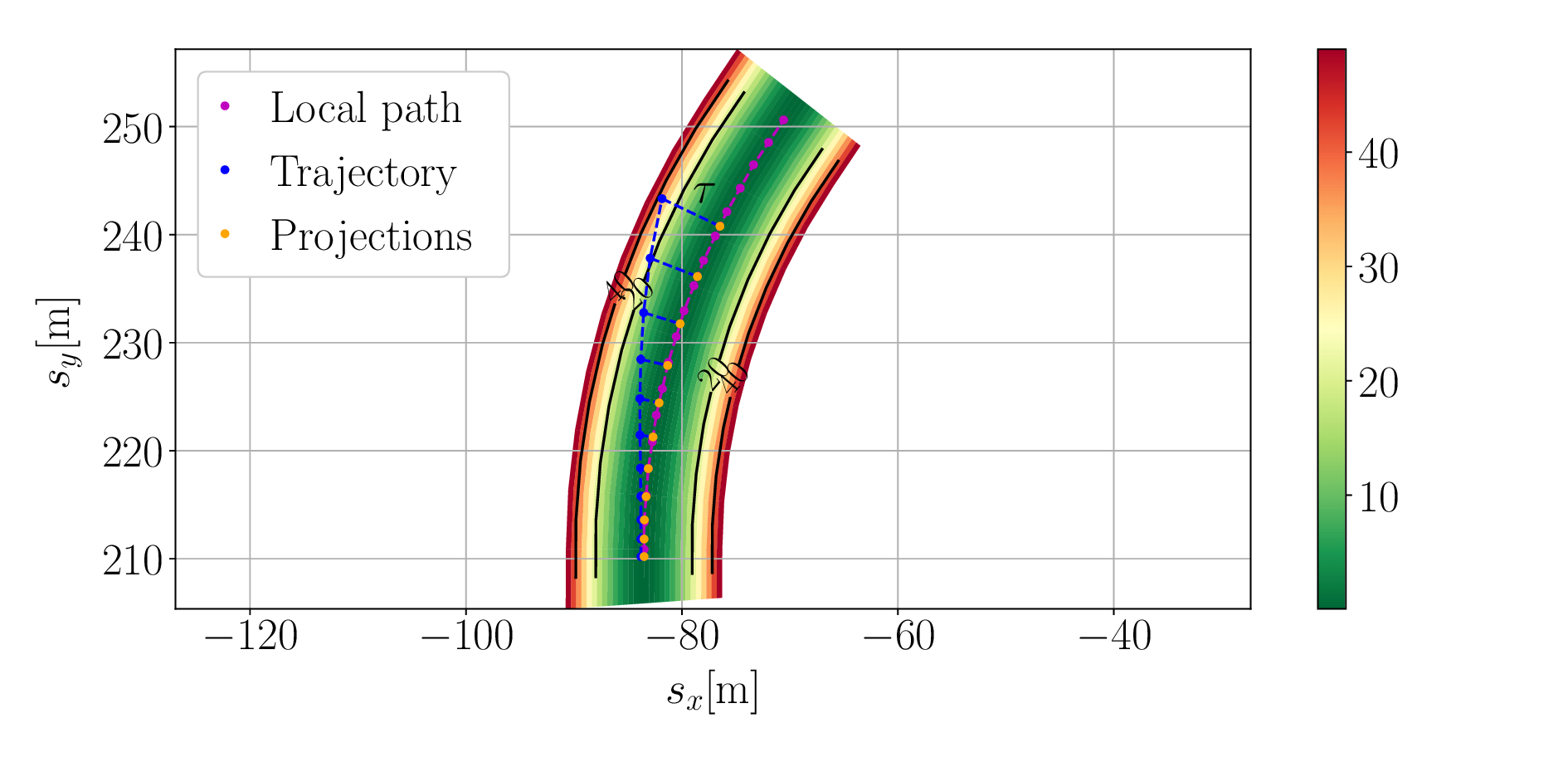}
    \vspace{-0.32cm}
    \caption{A scenario specific cost landscape is defined to determine the road related cost $c_4(X)$. Zero costs are assigned for the local path and the lateral offset $\boldsymbol{\tau}$ is penalized quadratically.}
    \label{fig:road_cost}
\end{figure}

Road data was obtained from the vehicle dynamics simulator CarMaker. The road environment is used for both evaluation scenarios (see Fig. \ref{fig:road_cost}). The kinematic single track model \cite{rajamani11} is used in this work. It describes the vehicle state $\boldsymbol{x}=[s_x, s_y, \delta, \upsilon, \psi]^\top \in \mathbb{R}^5$ via the global position $[s_x, s_y]$, the steering angle $\delta$, the longitudinal velocity $\upsilon$ and the orientation $\psi$. The control inputs $\boldsymbol{u}=[\upsilon_\delta, a] \in \mathbb{R}^2$ are the steering velocity $\upsilon_\delta$ and the longitudinal acceleration $a$. The Euler method with discretization time $\Delta t=0.1 \, \mathrm{s}$ is used for forward integration. With $N=80$ as the trajectory length, the optimization problem in \eqref{eq:MPC_problem} is solved over a time horizon of $T=8 \, \mathrm{s}$ which is inspired by the waymo open dataset challenge \cite{ettinger21}. This challenge deals with motion prediction which we consider as a future research topic.

\begin{figure}[tb]
    \vspace{-0.48cm}
    \includegraphics[width=0.39\textwidth]{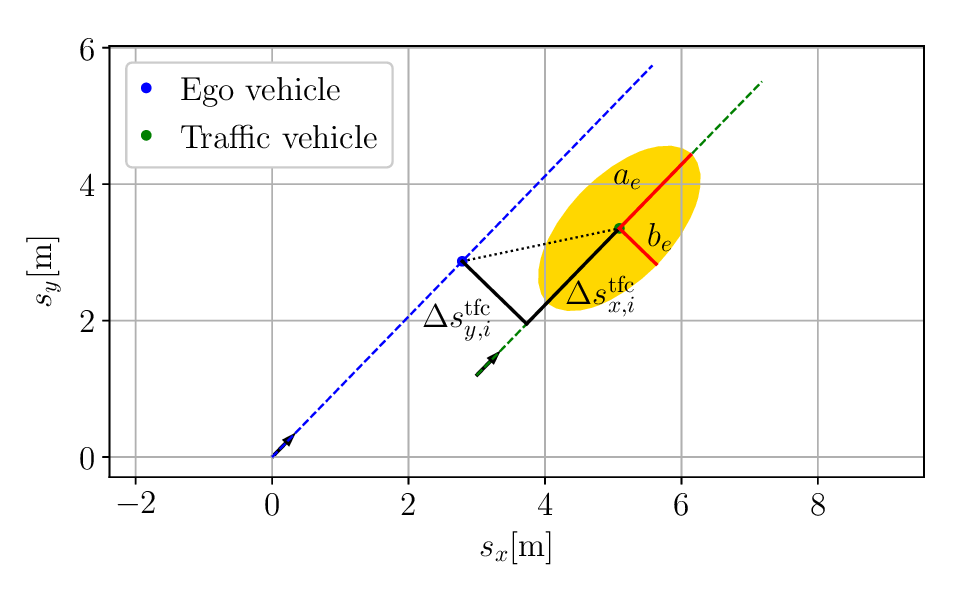}
    \vspace{-0.32cm}
    \caption{For the traffic cost $c_5(X)$, the distance between ego and traffic vehicle is computed in the coordinate system of the traffic vehicle. Longitudinal and lateral distances can be scaled with ellipsoidal parameters.}
    \label{fig:traffic}
\end{figure}

State $X \in \mathbb{R}^{5 \times N}$ and input trajectories $U \in \mathbb{R}^{2 \times N}$ are evaluated via the cost function $S(X,U)=\sum_{i=1}^5 \alpha_i c_i(\cdot)$ with the scaling parameters $\boldsymbol{\alpha} \in \mathbb{R}^5$. The cost terms are described in the following. The deviation from a desired veloctiy is penalized by the cost term $c_1(X)=\sum_{i=1}^N (\upsilon_i - \upsilon_\mathrm{des})^2$. The second cost term 
\begin{align}
    c_2(X) = \left\Vert \begin{pmatrix}
        s_{x,N} \\ s_{y,N}
    \end{pmatrix} - \begin{pmatrix}
        \bar{s}_{x,\mathrm{end}} \\\bar{s}_{y,\mathrm{end}}
    \end{pmatrix} \right\Vert_2
\end{align}
evaluates the offset between the final trajectory point $[s_{x,N}, s_{y,N}]$ and a desired end position $[\bar{s}_{x,\mathrm{end}}, \bar{s}_{y,\mathrm{end}}]$. The cost $c_3(U) = \sum_{i=1}^{N-1} (v_{\delta,i+1}-v_{\delta,i})^2 + (a_{i+1}-a_i)^2$
penalizes deviations in the control trajectories and thus has a regularizing effect. The road related cost $c_4(X) = \sum_{i=1}^N \tau_i^2$ encourages the vehicle to follow a local path $\boldsymbol{\gamma} \in \mathbb{R}^{2 \times m_\gamma}$ which is a set of reference positions (see Fig. \ref{fig:road_cost}). This cost term evaluates the lateral offsets $\tau \in \mathbb{R}^N$ between a spatial trajectory $\boldsymbol{s}_x, \boldsymbol{s}_y \in \mathbb{R}^N$ and the local path. The lateral offsets are computed 
\begin{align}
    \tau_i = \mathrm{project}(s_x^{(i)}, s_y^{(i)}, \boldsymbol{\gamma}) \quad \forall i \in \{1, N\}
\end{align}
by projecting every trajectory point on $\boldsymbol{\gamma}$. The fifth cost term aims to avoid collisions with other traffic participants and evaluates the distance between the ego vehicle's spatial trajectory and a traffic object (Fig. \ref{fig:traffic}). It is thereby assumed that the trajectory of a traffic participant with position and orientation $\boldsymbol{s}_x^{\mathrm{tfc}}, \boldsymbol{s}_y^{\mathrm{tfc}}, \boldsymbol{\psi}^{\mathrm{tfc}} \in \mathbb{R}^N$ is known over the full prediction horizon. For every prediction step, the position difference
\begin{align}
    \begin{bmatrix}
        \Delta s_{x,i}^\mathrm{tfc} \\ \Delta s_{y,i}^\mathrm{tfc}
    \end{bmatrix} = 
    \begin{bmatrix}
        \mathrm{cos}(\psi_i^\mathrm{tfc}) & \mathrm{sin}(\psi_i^\mathrm{tfc}) \\
        -\mathrm{sin}(\psi_i^\mathrm{tfc}) & \mathrm{cos}(\psi_i^\mathrm{tfc})
    \end{bmatrix} \cdot
    \begin{bmatrix}
        s_{x,i} - s_{x,i}^\mathrm{tfc} \\
        s_{y,i} - s_{y,i}^\mathrm{tfc}
    \end{bmatrix} 
\end{align}
is determined in the local coordinate system of the traffic vehicle. An ellipsoidal shape is introduced to compute a scaled distance
\begin{align}
    d_{e,i}&=\Bigl( \frac{\Delta s_{x,i}^\mathrm{tfc}}{a_e}\Bigr)^2 + \Bigl( \frac{\Delta s_{y,i}^\mathrm{tfc}}{b_e}\Bigr)^2.
\end{align}
with width $a_e=6\, \mathrm{m}$ and $b_e=2\, \mathrm{m}$. These parameters allow to assign different weights to longitudinal and lateral distances. The traffic related cost results as $c_5(X)=\sum_{i=1}^N \frac{1}{d_{e,i}^2}$. In both scenarios, the same cost functions are used and the performance of the algorithm is evaluated for every sampling approach. The covariance matrix $\boldsymbol{\Sigma}=\mathrm{diag}(0.1, 2)$ is selected for the BG while $\boldsymbol{\Sigma}=\mathrm{diag}(0.045, 1.1)$ is used in the IL approach. The matrices $\boldsymbol{\Sigma}^{(1)}=\mathrm{diag}(0.03, 0.075)$ and $\boldsymbol{\Sigma}^{(2)}=\mathrm{diag}(0.045, 0.09)$ are chosen for the 2DF sampling. The number of trajectories samples is $K=200$ for every sampling distribution. The MPPI weighted averaging scheme from \eqref{eq:weight_determination} uses $\lambda=5$. The scaling parameters for $S(X, U)$ are chosen as $\boldsymbol{\alpha}=[0.5, 10, 0.06, 1, 4.5]^\top$.

\subsection{Normalizing flow training}

\begin{figure}
    \centering
    \includegraphics[width=0.41\textwidth]{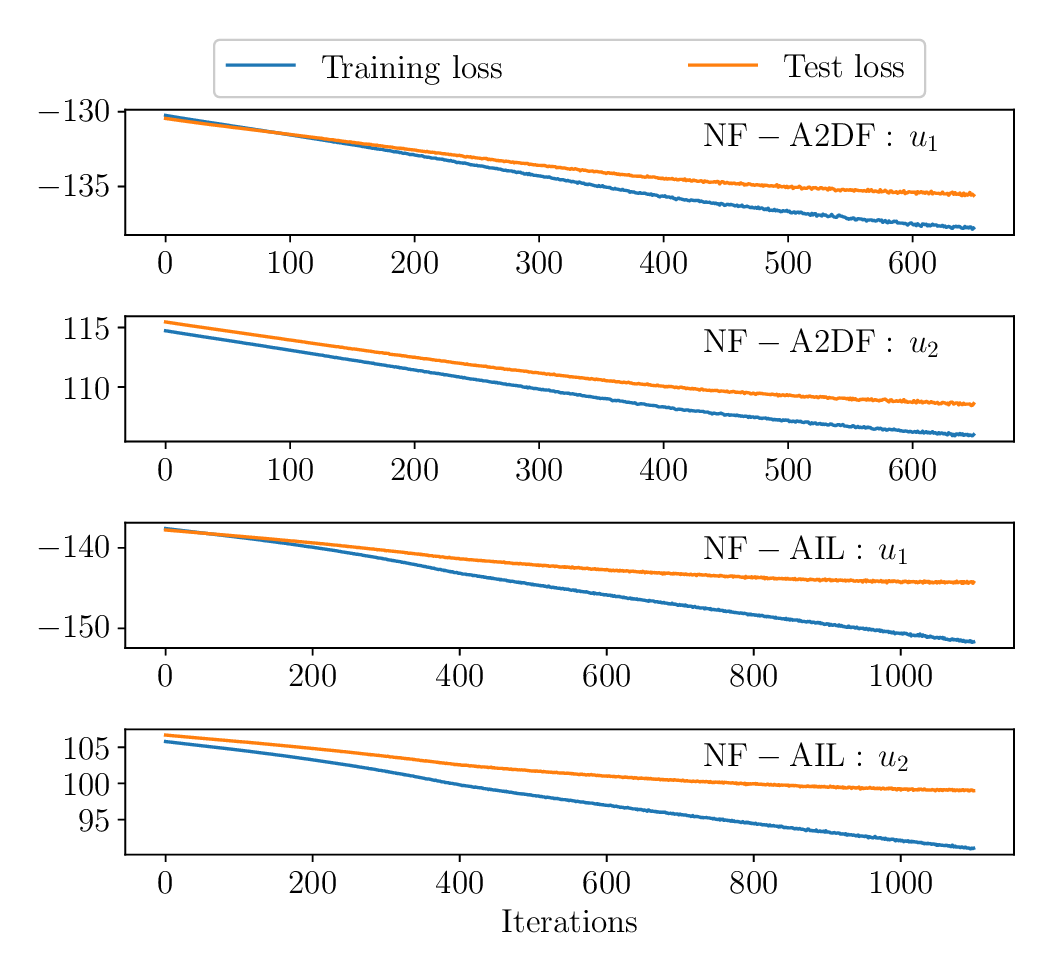}
    \vspace{-0.32cm}
    \caption{Loss curves for model training. When the test loss can no longer be decreased, the training is terminated.}
    \label{fig:loss_curves}
\end{figure}
In the following, the training processes for the NF-A2DOF and NF-AIL approach are described. The trained models are later used for the sampling of input trajectories. Hereby, separate flow models for $u_1=\upsilon_\delta$ and $u_2=a$ are trained. The training process for the NF-A2DOF sampling approach is described first. Training trajectories are generated via the steps described in Alg. 1. Hereby, $\epsilon_{\mathrm{draw}}^{(1)}=0.03$ is used for $u_1$ while $\epsilon_{\mathrm{draw}}^{(2)}=0.9$ is selected for $u_2$. In both cases, the number of training trajectories is $B=400$ and the switching variance is chosen as $\epsilon_{\mathrm{switch}} = 220$. The generated trajectories are split into training and test data with a $60/40$ ratio. The loss curves for the training are given in Fig. \ref{fig:loss_curves}. In order to prevent overfitting, the training process is stopped when the loss on the test data can no longer be decreased. As a consequence, both models are trained for $650$ steps.

The training samples for the NF-AIL approach are generated via the steps in Alg. 2. Again, a separate model is trained for each input. The drawing variances are $\epsilon_\mathrm{draw}=0.045$ for $u_1$ and $\epsilon_\mathrm{draw}=1.1$ for $u_2$. The switching variance is selected as $\epsilon_{\mathrm{switch}} = 350$ and the number of trajectories is $B=400$ for both cases. The trajectories are again divided into training and test data with a $60/40$ split. Figure \ref{fig:loss_curves} shows the loss curves for the model training. The model training is ended when the loss on the test data increases again which happens after $1100$ steps.

\subsection{Scenario with static traffic objects}

\begin{figure}
    \includegraphics[width=0.46\textwidth]{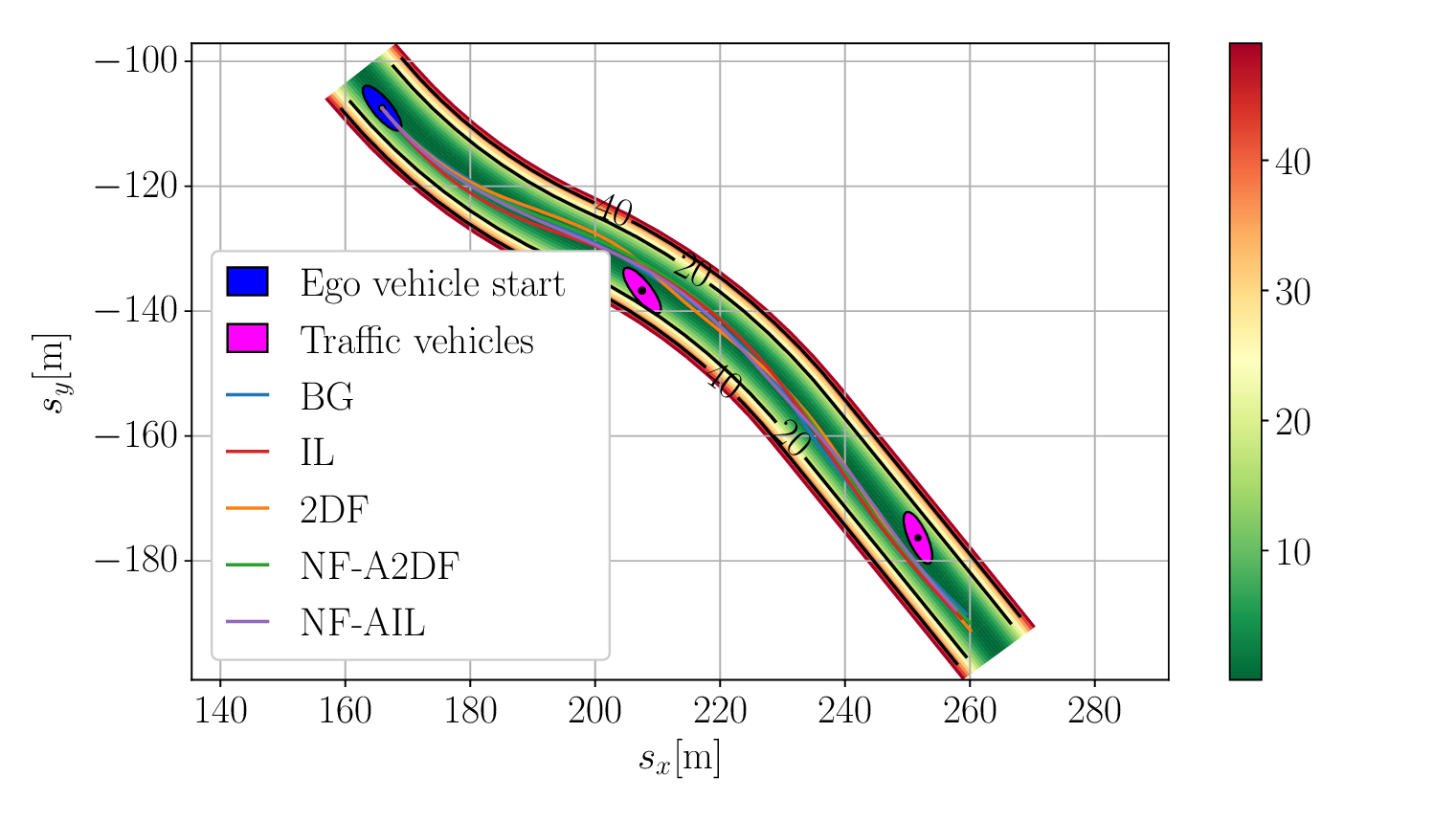}
    \vspace{-0.32cm}
    \caption{Spatial trajectories for the static traffic scenario with desired velocity $\upsilon_\mathrm{des}=6\frac{\mathrm{m}}{\mathrm{s}}$ and total duration $T_\mathrm{end}=45\, \mathrm{s}$.}
    \label{fig:spatial_trajectories}
\end{figure}

The first driving scenario features four static traffic vehicles and has a total driving distance of $250 \, \mathrm{m}$. The ego vehicle starts from the initial velocity $\upsilon_0 = 0 \, \frac{\mathrm{m}}{\mathrm{s}}$. Three variants of the scenario are created and a different value for the desired velocity $\upsilon_\mathrm{des}$ is selected in every variant. Ten testruns are performed for every variant. Figure \ref{fig:spatial_trajectories} shows spatial trajectories for one testrun with $\upsilon_\mathrm{des} = 6 \, \frac{\mathrm{m}}{\mathrm{s}}$. The average planning costs and the relative cost reduction compared to the BG approach are given in Tab. \ref{tab:costs_scenario_1}. While all sampling distributions generate similar spatial profiles, considerable differences occur in terms of planning costs. It can be seen that sampling with the IL and 2DF approach leads to a substantial reduction in planning costs compared to the BG. It can further be observed that the normalizing flow based approaches achieve the highest overall performance for every variant.

\begin{table}[htbp]
    \caption{Planning cost for scenario with static traffic with desired velocity $\upsilon_\mathrm{des}$ and total duration $T_\mathrm{end}$.}
    \vspace{-0.2cm}
    \scriptsize
    \centering
    \begin{tabular}{|c|C{0.92cm}|C{0.92cm}|C{0.92cm}|C{1.1cm}|C{0.92cm}|}
        \hline 
        \multicolumn{1}{|c|}{\raisebox{2.5ex}} &\multicolumn{5}{c|}{Variant 1: $\upsilon_\mathrm{des}=6 \, \frac{\mathrm{m}}{\mathrm{s}}$,
        $T_\mathrm{end}=45\, \mathrm{s}$} \\[2pt]
        \cline{2-6}
         \multicolumn{1}{|c|}{\raisebox{2ex}} & BG & IL & 2DF & NF-A2DF & NF-AIL \\[-1pt]
         \hline
         $c_1$ &87.0	&63.7	&59.5	&53.5	&45.6 \\
        $c_2$ &33.8	&32.7	&31.3	&29.3	&26.8 \\
        $c_3$ &64.4	&0.2	&13.4	&8.6	&0.1 \\
        $c_4$ & 119.0&	117.8	&110.1	&101.7	&103.3 \\ 
        $c_5$ &184.4	&73.1	&55.0	&57.9	&59.9 \\ 
        $S(X,U)$ &488.6	&287.5&269.3 &251.0 &235.7 \\ 
        & & (-41\%)	&(-45\%)	&(-49\%)	&(-52\%)\\ 
        \hline \hline
        \multicolumn{1}{|c|}{\raisebox{2.5ex}} &\multicolumn{5}{c|}{Variant 2: $\upsilon_\mathrm{des}=8 \, \frac{\mathrm{m}}{\mathrm{s}}$,
        $T_\mathrm{end}=35\, \mathrm{s}$} \\[2pt]
        \cline{2-6}
        \multicolumn{1}{|c|}{\raisebox{2ex}} & BG & IL & 2DF & NF-A2DF & NF-AIL \\[-1pt]
        \hline
        $c_1$ &159.8	&113.9	&127.6	&105.7	&95.1 \\
        $c_2$ &49.0	&46.8	&46.8	&43.1	&40.8 \\
        $c_3$ &67.0	&0.2	&13.8	&9.1	&0.1 \\
        $c_4$ &199.2	&160.4	&161.0	&146.4	&142.1 \\
        $c_5$ &64.3	&83.7	&70.9	&64.0	&75.4 \\
        $S(X,U)$ &539.3	&404.9	&420.1	&368.2	&353.5 \\
        & &(-25\%)	&(-23\%)	&(-32\%)	&(-35\%)\\
        \hline \hline
        \multicolumn{1}{|c|}{\raisebox{2.5ex}} &\multicolumn{5}{c|}{Variant 3: $\upsilon_\mathrm{des}=10 \, \frac{\mathrm{m}}{\mathrm{s}}$,
        $T_\mathrm{end}=28\, \mathrm{s}$} \\[2pt]
        \cline{2-6}
        \multicolumn{1}{|c|}{\raisebox{2ex}} & BG & IL & 2DF & NF-A2DF & NF-AIL \\[-1pt]
        \hline
        $c_1$ &275.3	&201.6	&201.5	&178.11	&204.1\\
        $c_2$ &70.5	&73.2	&68.1	&62.4	&65.6\\
        $c_3$ &68.0	&0.2	&14.2	&9.4	&0.1\\
        $c_4$ &287.5	&268.6	&243.9	&244.7	&206.8\\
        $c_5$ &73.8	&92.1	&125.0	&87.6	&107.6\\
        $S(X,U)$ &775.2	&635.6	&652.8	&582.2	&584.3\\
         & &(-19\%)	&(-16\%)	&(-25\%) &(-25\%)\\
        \hline
    \end{tabular}
    \label{tab:costs_scenario_1}
\end{table}

\begin{table}[htbp]
    \scriptsize
    \centering
    \caption{Planning cost for scenario with dynamic traffic with velocities $\upsilon_\mathrm{des}$, $\boldsymbol{\upsilon}^{\mathrm{tfc}}$ and total duration $T_\mathrm{end}$.}
    \vspace{-0.2cm}
    \begin{tabular}{|c|C{0.92cm}|C{0.92cm}|C{0.92cm}|C{1.1cm}|C{0.92cm}|}
        \hline
        \multicolumn{1}{|c|}{\raisebox{2.5ex}} &\multicolumn{5}{c|}{Variant 1: $\upsilon_\mathrm{des}=8\frac{\mathrm{m}}{\mathrm{s}}$, \, $\boldsymbol{\upsilon}^{\mathrm{tfc}}=[4, 5] \, \frac{\mathrm{m}}{\mathrm{s}}$, $T_\mathrm{end}=30\, \mathrm{s}$} \\[2pt]
        \cline{2-6}
         \multicolumn{1}{|c|}{\raisebox{2ex}} & BG & IL & 2DF & NF-A2DF & NF-AIL \\[-1pt]
         \hline
            $c_1$ &166.4	&121.11	&127.0	&116.5	&126.1\\
            $c_2$ &51.3	&49.0	&48.8	&44.1	&45.3\\
            $c_3$ &65.9	&0.2	&13.3	&8.7	&0.1\\
            $c_4$ &187.5	&158.4	&160.7	&143.6	&120.9\\
            $c_5$ &70.9	&85.6	&68.9	&72.7	&88.9\\
            $S(X,U)$ &541.9	&414.3	&418.7	&385.7	&381.3\\
            & &(-24\%)	&(-23\%)	&(-27\%)	&(-30\%)\\
        \hline \hline
        \multicolumn{1}{|c|}{\raisebox{2.5ex}} &\multicolumn{5}{c|}{Variant 2: $\upsilon_\mathrm{des}=10\frac{\mathrm{m}}{\mathrm{s}}$, $\boldsymbol{\upsilon}^{\mathrm{tfc}}=[4, 5] \, \frac{\mathrm{m}}{\mathrm{s}}$, 
    $T_\mathrm{end}=18\, \mathrm{s}$} \\[2pt]
        \cline{2-6}
        \multicolumn{1}{|c|}{\raisebox{2ex}} & BG & IL & 2DF & NF-A2DF & NF-AIL \\[-1pt]
        \hline
        $c_1$ &445.0	&232.3	&207.2	&245.1	&193.7\\
             $c_2$ &97.5	&71.8	&69.9	&77.1	&69.0\\
             $c_3$ &68.6	&0.2	&14.2	&9.12	&0.1\\
             $c_4$ &545.7	&252.5	&205.9	&160.6	&132.0\\
             $c_5$ &85.6	&100.6	&123.7	&91.6	&131.5\\
             $S(X,U)$ &1242.4	&657.4	&620.9	&583.6	&526.2\\
             & &(-47\%)	&(-50\%)	&(-53\%)	&(-57)\%\\
        \hline \hline
        \multicolumn{1}{|c|}{\raisebox{2.5ex}} &\multicolumn{5}{c|}{Variant 3: $\upsilon_\mathrm{des}=10\frac{\mathrm{m}}{\mathrm{s}}$, $\boldsymbol{\upsilon}^{\mathrm{tfc}}=[5, 6] \, \frac{\mathrm{m}}{\mathrm{s}}$, $T_\mathrm{end}=18\, \mathrm{s}$} \\[2pt]
        \cline{2-6}
        \multicolumn{1}{|c|}{\raisebox{2ex}} & BG & IL & 2DF & NF-A2DF & NF-AIL \\[-1pt]
        \hline
        $c_1$ &353.2	&265.0	&286.4	&208.4	&255.4\\
             $c_2$ &178.3	&143.0	&120.8	&120.8	&121.3\\
             $c_3$ &66.6	&0.2	&14.0	&9.3	&0.1\\
             $c_4$ &312.9	&255.0	&261.6	&228.5	&186.5\\
             $c_5$ &93.6	&85.6	&84.5	&72.0	&77.0\\
             $S(X,U)$ &1004.6	&748.7	&767.3	&638.9	&640.3\\
               &&(-25\%)	&(-23\%)	&(-36\%)	&(-36\%)\\
        \hline     
    \end{tabular}
    \label{tab:costs_scenario_2}
\end{table}

\subsection{Scenario with moving traffic objects}

The second scenario features two dynamic traffic vehicles. Collision avoidance is thus harder then in the first scenario. The total driving distance is $130 \, \mathrm{m}$ and the ego vehicle again starts from $\upsilon_0 = 0 \, \frac{\mathrm{m}}{\mathrm{s}}$. Three variants for the scenario are created. Different values for the desired velocity $\upsilon_\mathrm{des}$ and the traffic velocities $\boldsymbol{\upsilon}^{\mathrm{tfc}}$ are selected in every variant. Ten testruns are carried out for every variant. Note that the traffic vehicles have constant lateral position and velocity. The average planning costs and the relative cost reduction compared to the BG approach are given in Tab. \ref{tab:costs_scenario_2}. For every variant of the scenario, sampling via the IL and 2DF approach results in a considerable improvement over the BG. The overall lowest planning costs are again achieved by the normalizing flow based approaches.
\section{Conclusion}
\label{sec:conclusion}

In this paper, a MPPI based trajectory planning algorithm is proposed. Therefore, several approaches for the stochastic sampling of input trajectories are considered. In addition to simple sampling concepts, normalizing flows are introduced to achieve a more efficient exploration of the input domain. In the training process of normalizing flows, probabilistic rules are utilized to adapt trajectories that are generated by means of existing approaches. The adapted trajectories therefore construct a distribution which is learned by a normalizing flow. The performance of the trajectory planning algorithm is analyzed in two simulation scenarios for all sampling concepts.
In both scenarios, a considerable performance improvement is observed for the normalizing flow based sampling approaches. Future research could be the incorporation of motion prediction of other road users. Additionally, the proposed methodology should be evaluated in different driving scenarios and with real-world datasets to analyze its generalizability in more real world settings.

\end{document}